\def\year{2021}\relax
\documentclass[letterpaper]{article} 
\usepackage{aaai21}  
\usepackage{times}  
\usepackage{helvet} 
\usepackage{courier}  
\usepackage[hyphens]{url}  
\usepackage{graphicx} 
\usepackage{natbib}  
\usepackage{caption} 
\usepackage{multirow}
\usepackage{gensymb}
\usepackage{subfigure}

\title{Sketch Generation with Drawing Process Guided by Vector Flow and Grayscale}

\author {
    Zhengyan Tong,\textsuperscript{\rm 1}
    Xuanhong Chen,\textsuperscript{\rm 1,2} 
    Bingbing Ni,\textsuperscript{\rm 1,2}\thanks{Corresponding author: Bingbing Ni.} 
    Xiaohang Wang \textsuperscript{\rm 1} \\
}
\affiliations {
    \textsuperscript{\rm 1} Shanghai Jiao Tong University\\
    \textsuperscript{\rm 2} Huawei Hisilicon \\
    \{418004, chen19910528, nibingbing, xygz2014010003\}@sjtu.edu.cn
}

\begin{document}

\maketitle

\begin{abstract}
We propose a novel image-to-pencil translation method that could not only generate high-quality pencil sketches but also offer the drawing process. Existing pencil sketch algorithms are based on texture rendering rather than the direct imitation of strokes, making them unable to show the drawing process but only a final result. To address this challenge, we first establish a pencil stroke imitation mechanism. Next, we develop a framework with three branches to guide stroke drawing: the first branch guides the direction of the strokes, the second branch determines the shade of the strokes, and the third branch enhances the details further. Under this framework's guidance, we can produce a pencil sketch by drawing one stroke every time. Our method is fully interpretable. Comparison with existing pencil drawing algorithms shows that our method is superior to others in terms of texture quality, style, and user evaluation. Our code and supplementary material are now available at: \url{https://github.com/TZYSJTU/Sketch-Generation-with-Drawing-Process-Guided-by-Vector-Flow-and-Grayscale}
\end{abstract}

\begin{figure*}
  \includegraphics[width=\textwidth]{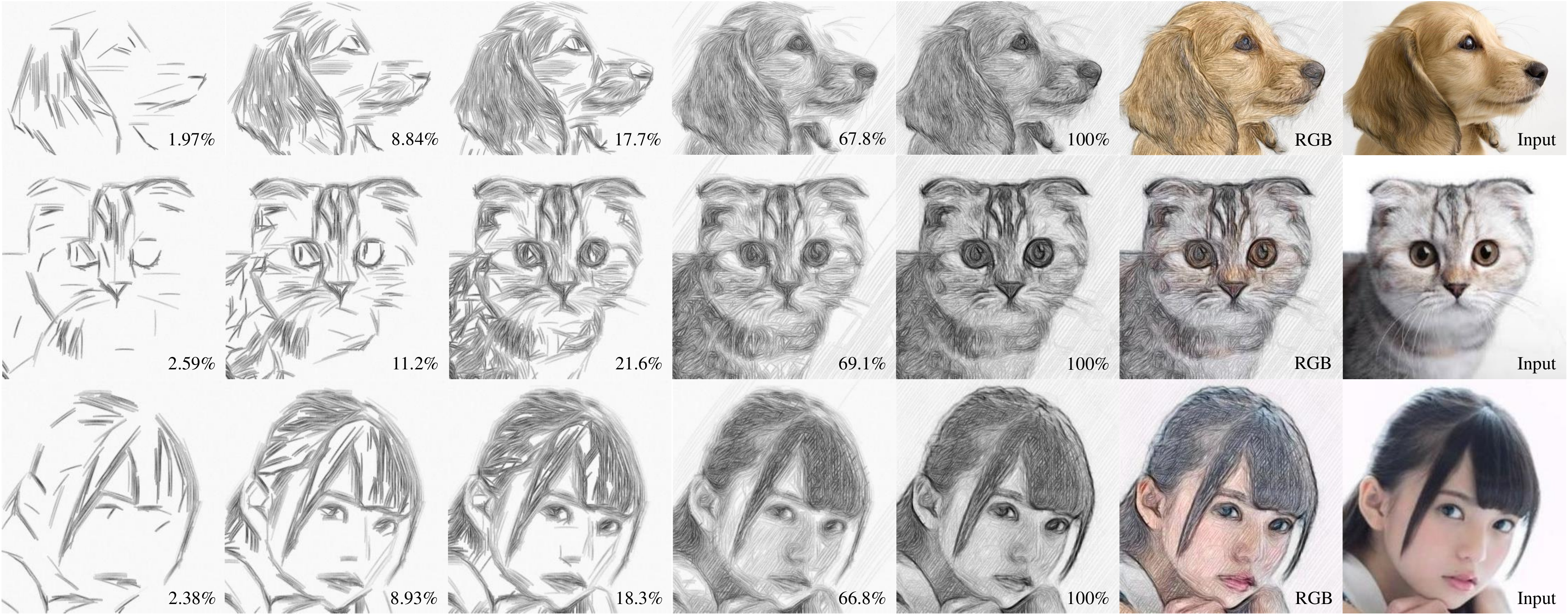}
  \caption{Given a natural image (in the rightmost column) as input, our algorithm can produce a pencil sketch with the process by drawing one stroke every time. For the dog, the cat, and the girl, we draw 10390, 11529, and 16698 strokes. The percentage in the lower right corner of each picture is the proportion of the number of strokes that have been drawn in this picture to the number of strokes in the final result.}
  \label{process}
\end{figure*}

\section{Introduction}
The pencil sketch is one kind of drawing with a highly realistic style. It is not only a popular form of art creation but also the essential basic of other art forms such as oil painting. To draw a pencil sketch, artists should have accurate contour configuration ability and superb shading drawing skills, requiring long-term professional training. Therefore, there has always been a strong demand for the pencil sketch rendering algorithm.

The existing pencil drawing algorithms are mainly implemented by Non-Photorealistic Rendering (NPR) \cite{rosin2012image} and can be further divided into 3D model-based sketching and 2D image-based sketching \cite{lu2012combining}. 3D models can provide the complete geometric information of the objects, and the lighting condition in the 3D scene is fully controllable. Thus 3D model-based sketching can accurately grasp the spatial structure and render the shading texture according to the light condition. However, in most application scenarios, we can not get 3D models but only 2D natural images, so there is a greater demand for image-based sketch rendering algorithms. For 2D natural images, the geometric information is often incomplete, and the light components are usually complicated and noisy, making it hard to do sketch rendering. Generally, mathematical rendering algorithms can maintain the structures well, control the rendering effect in a fine-grained manner, and are often interpretable. In recent years, many deep learning methods for image style transfer tasks have been developed. Results of these methods usually have a stronger style than those of procedural rendering algorithms. However, due to the complexity of pencil drawing, neural methods do not perform well at capturing pencil sketch texture \cite{li2019im2pencil}. Deep learning style translation usually suffers structure distortion and artifacts, which is a serious defect for the pencil drawing task that requires reserving structures and producing high-quality textures. Besides, neural networks' parameter control mechanisms are usually high-level and are difficult to explain. 

Whether it is a procedural algorithm or a deep learning neural style transferer, existing algorithms can only get the final result without offering the drawing process. Actually, there exists very limited research on auto-drawing with the drawing process. Our algorithm implements image-to-pencil with a drawing process for the first time, which significantly enhances our algorithm's novelty. As shown in Figure \ref{process}, given an image (in the rightmost column), our algorithm can produce a pencil sketch by drawing one stroke at a time (Figure \ref{process} only shows the drawing process in stages, the whole process can be found in the supplementary material). Our algorithm could generate high-quality details, and the final result has a strong style. Besides, we can produce pencil drawings with obviously different visual effects by adjusting the strokes' properties. Comparison with existing pencil drawing algorithms shows that our method performs favorably in terms of texture quality, style, and user evaluation. 


Our work is inspired by the observation of real pencil drawings, and our algorithm models the real artists' drawing techniques. Thus the interpretability of our method is stronger than others. Since pencil drawing has many different drawing styles and artists use various drawing tools \cite{dodson1990keys}, we only simulate the most popular sketching method. That is, drawing strokes with diverse directions, shades, lengths, and widths on a canvas to gradually form a picture. For the strokes' direction, artists usually use the tangent direction of objects' edges/contours to guide the strokes' direction; for the strokes' shade, artists usually adjust their pencil sketches' contrast higher than real light conditions to make their works more visually impactful \cite{kelen1974leonardo, hoffman1989vision}. We divide the pencil sketching task into two steps. For the first step, we develop a parameter-controlled pencil stroke generation mechanism based on the pixel-scale statistical results of some real pencil drawings. For the second step, we develop a framework to guide strokes arranging on the canvas. Finally, we implement the pencil sketch auto-drawing technique.

In this work, our main contribution is that we propose a novel image-to-pencil translation method that could generate high-quality results and offer the drawing process. 

\section{Related Work}

\subsection{Non-Photorealistic Rendering}
There is a rich research history on the non-photorealistic texture rendering of pencil drawing.
3D models provide all the geometric information and light conditions, thus convenient for pencil drawing rendering. \cite{lake2000stylized} presented pencil sketching texture mapping technique and proposed pencil shading rendering. \cite{lee2006real} detected contours from 3D models and imitated human contour drawing. For expression of shading, they mapped oriented textures onto objects' surface. \cite{praun2001real} achieved real-time hatching rendering over arbitrary complex surfaces using 3D models. These 3D model-based methods usually obtain satisfactory results. However, when the 3D structure and light conditions are not available, these methods cannot work.

2D image-based methods mainly include the following typical algorithms. Sousa and Buchanan presented an observational model of blenders and kneaded eraser \cite{sousa1999observational}, and simulated artists and illustrators' graphite pencil rendering techniques in \cite{sousa1999computer}. \cite{chen2004example} proposed a composite sketching approach for portrait drawing. \cite{durand2001decoupling} presented an interactive system that allowed users to produce drawings in a variety of styles, including pencil sketching. \cite{mao2002automatic} detected the input image's local structure orientation and adopted linear integral convolution (LIC) to render sketch texture. \cite{yamamoto2004enhanced} divided the input image into several layers of successive intensity ranges, then did rendering for each layer and finally added them together. \cite{li2003feature} analyzed the image moment and texture of each region, using the captured feature geometric attributes to implement pencil drawing rendering. Others proposed some improved LIC-based method \cite{chen2008novel, gao2010automatic, kong2018hybrid, chen2017stylebank}. Pencil sketch rendering could also be implemented by image analogies \cite{hertzmann2001image}. \cite{lu2012combining} proposed a novel two-stage system combining both line and tone for pencil drawing production and obtained significantly better effect than the above methods.  

\subsection{Drawing with Process}
All previous works on the image-to-pencil task can only generate a final result, without offering the drawing process. Here we review some related Stroke-Based Rendering methods which have a process. 

\cite{fu2011animated} proposed an algorithm that used human pre-drawn line drawing as the input to derive a stroke order and animate the sketching automatically, but this method couldn't well recover the input line drawing. 
\cite{ha2018a} presented a RNN-based method trained on a dataset of human-drawn simple images to draw stick figures.
StrokeNet \cite{zheng2018strokenet} can generate a sequence of only a few strokes to write Chinese characters. However, the strokes as well as their sequence are very different from that of human writing.
\cite{huang2019learning} adopted model-based Deep Deterministic Policy Gradient (DDPG) algorithm to train a neural agent to learn to do oil painting with process. However, this method cannot be directly applied to pencil drawing because pencil strokes' characteristics and fusion mode are distinctive from the oil painting. The strokes of pencil drawing are lines while the oil painting' strokes are color blocks; the newly drawn strokes cannot cover the old ones in pencil drawing while the oil painting's strokes can. Besides, lines' sparsity makes it hard to train pencil drawing neural agents. Deep reinforcement learning (DRL) requires a massive amount of parameters when training, so the network's input size is very limited. \cite{huang2019learning}'s oil agent can only handle $128 \times 128$ images, unable to generate fine-grained details, while our algorithm has no restriction on the size of the input image and could generate high-quality details.

\begin{figure}
  \includegraphics[width=\linewidth]{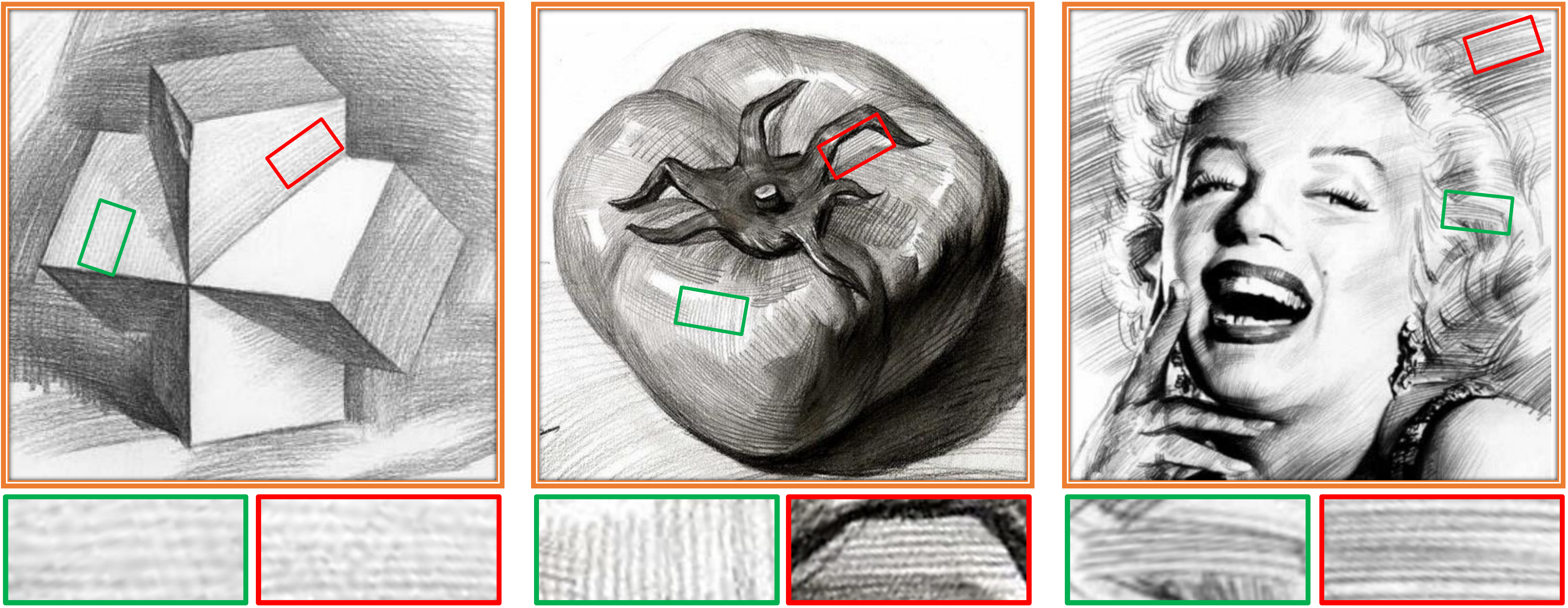}
  \caption{Three real pencil drawings. It can be seen from the zoomed-in areas that the texture of pencil drawings is actually some parallel strokes.}
  \label{real}
\end{figure}
\section{Stroke Simulation}
Lines are the fundamental elements of pencil sketching. 
Since pencil drawing strokes are lines, we regard the “line" and the “stroke" as the same concept in this article. 

\subsection{Observation and Statistic}
The analysis of real pencil drawings can be performed globally or locally. The statistics of global features are mainly on the histogram. \cite{lu2012combining} counted and fitted the histogram distribution of several real pencil drawings, then did histogram matching to transfer the tone of input images. The analysis of local features is mainly for texture. \cite{sousa1999observational} observed the blenders and erasers' absorptive and dispersive properties, and studied their interacting with lead material which deposited over the drawing paper. \cite{hertzmann2000illustrating} analyzed different hatching styles of pencil sketch. \cite{xie2007efficient} studied the graphite's distribution in pencil drawings and made three assumptions about the local distribution characteristics. Generally, local features are more important than global features because local features can better reflect the characteristics of pencil drawings, while the global histogram distribution of various artists' works is often personalized. Our observation and analysis methods are entirely based on local features.

\begin{figure}[h]
    \includegraphics[width=\linewidth]{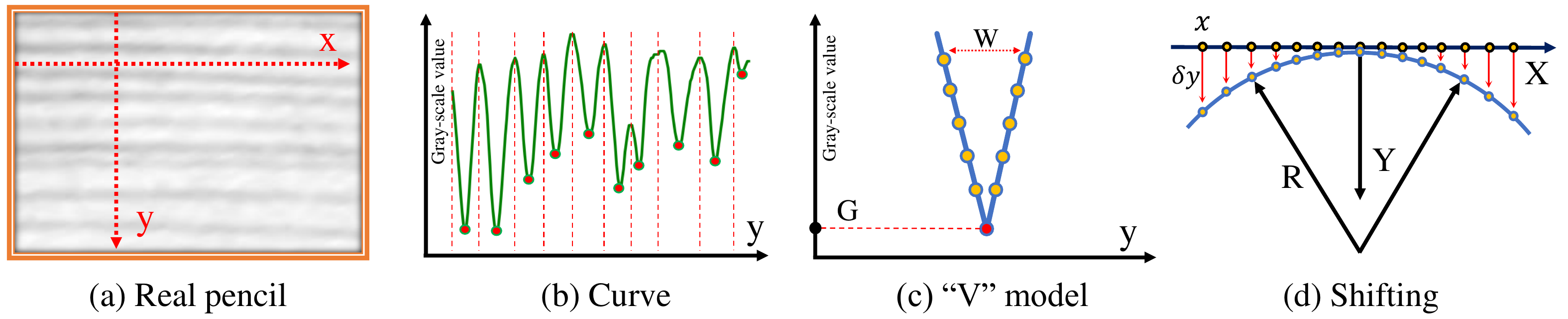}
	\caption{(a) is a patch cut from a real pencil drawing; (b) shows the gray value of the pixels on the red dotted line marked with \emph{y} in (a); (c) is the fitting of the gray value's periodic change in (b). (d) is the illustration of stroke bending.} 
	\label{patch}
\end{figure}

Three real pencil drawings are shown in Figure \ref{real}. It can be seen from the zoomed-in area that the texture is composed of many parallel curves. This pattern is also prevalent and evident in the rest region of these drawings. For any group of parallel curves, lines within the group have a high degree of similarity. That is, the distance between any two adjacent lines, the shade, length, and width of each line are very close. Therefore, we can perform statistical analysis on these parallel lines. In order to facilitate statistics, we cut some patches just containing one set of parallel curves from some realistic pencil drawings, and then rotate the patches until the lines' direction is nearly horizontal, as shown in Figure \ref{patch}(a). We notice that the lines are slightly curved and not strictly parallel, which brings great difficulty to the statistics in the horizontal direction, as shown by the red dotted line marked with \emph{x} in Figure \ref{patch}(a). However, the lines' curving does not affect the gray values' distribution in the vertical direction, as shown by the red dotted line marked with \emph{y} in Figure \ref{patch}(a). So we do statistics in the vertical direction. For the pixels on the red dotted line marked with \emph{y} in Figure \ref{patch}(a), their gray values are shown in Figure \ref{patch}(b). We draw some red dotted lines at all peak points in (b), it shows that the gray value's distribution curve between any two adjacent peak points is close to the letter V, as shown in Figure \ref{patch}(c). In fact, each “V" curve in Figure \ref{patch}(b) corresponds to a stroke in Figure \ref{patch}(a). The same statistics are performed on all the columns of pixels. Suppose there are \emph{n} columns, then each stroke corresponds to \emph{n} “V". For the \emph{n} “V" corresponding to a particular stroke, the gray value of pixels at the same positions (the yellow points in Figure \ref{patch}(c)) can be assumed to be independent and identically distributed. Thus, the gray value's mean and variance of each pixel in each “V" can be calculated.

We use the following method for fitting. We define two variables to determine a “V" curve: width \emph{W} and the mean value \emph{G} of the central pixel's gray value. As shown in Figure \ref{patch}(c), \emph{W} represents the pixel amount of the “V" curve. The red dot indicates the central pixel of the “V" curve. \emph{G} is the mean value of the central pixel's gray value. Assume a pixel on this curve is \emph{d} pixels away from the central red pixel, then the gray value's mean and variance of this pixel can be calculated by the following equations:
\begin{equation}
    mean(d)=G+(255-G)\times\frac{2d}{W-1}
\end{equation}
\begin{equation}
    variance(d)=(255-G)\times\cos{\frac{\pi d}{W-1}}
\end{equation}
Now suppose we can straighten the lines in Figure \ref{patch}(a) in the horizontal direction, and assuming the length of the line is \emph{L}, then these lines can be represented by a gray value matrix with \emph{W} rows and \emph{L} columns. For the pixels in one specific row, their gray value can be considered to be independent and identically Gaussian distributed. Now we define a matrix \emph{F} with the shape of $(W, 2)$ to record the gray value's distribution of a line. Element $(w, 1)$ and $(w, 2)$ in \emph{F} indicates the gray value's mean and variance of all the pixels in row \emph{w} in the line. As long as \emph{G} and $W$ are specified, the distribution matrix \emph{F} of the line can be calculated.

\subsection{Stroke Generation}
\begin{figure}[h]
    \includegraphics[width=\linewidth]{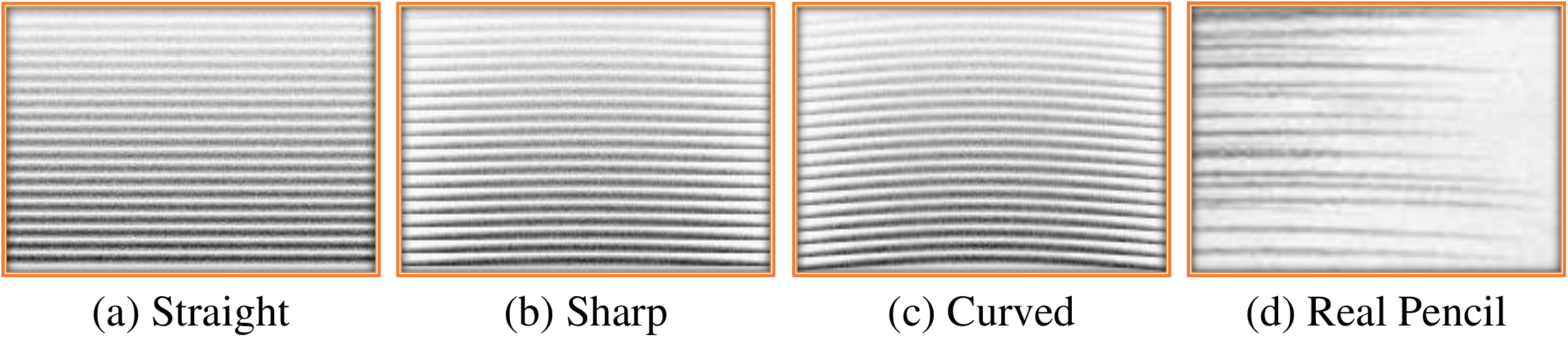}

	\caption{Stroke generation steps} 
	\label{imitation}
\end{figure}

We first simulate a straight line and then bend it in the vertical direction to get a more natural effect. To draw a straight line, we need to specify three parameters: central pixel's gray value mean \emph{G}, line width \emph{W}, and line length \emph{L}. Firstly, use \emph{G} and \emph{W} to calculate the distribution matrix \emph{F} of this line. Then for every row, the gray value of each pixel is randomly generated according to \emph{F}. As shown in Figure \ref{imitation}(a), We have drawn some lines with the width $W=7$ pixels but with different \emph{G}. These straight lines look too rigid, so we adjust their shape further.
By observing realistic pencil drawing strokes, as shown in Figure \ref{imitation}(d), we found that the head/tail of the strokes are thinner and lighter than the middle part, which is because when the pencil tip just touches the paper's surface or when it is about to leave, the pressure of the pencil tip on the paper's surface is less than when drawing the middle part of the line. Lines are not entirely straight but are slightly curved is because artists draw lines by swinging their wrists, so the movement of the pencil tip on paper is essentially a circular motion with a large radius. We bend the previously generated straight lines twice to achieve these effects. For the first time of bending, as shown in Figure \ref{patch}(d), the yellow dots on the X-axis indicate the pixels in a particular row of the line. These pixels will be shifted to the blue circle. Use the midpoint of the line as the origin to establish a coordinate system, pixels with different abscissas on the line will have different degrees of deviation in the Y direction. Assuming the blue circle radius is \emph{R}, pixels with abscissa \emph{x} will be shifted by $\delta y(x)$ pixels in the Y direction. The radius \emph{R} and offset $\delta y(x)$ can be calculated as $R=\frac{L^{2}}{4W}$ and $\delta y(x)=\frac{x^{2}}{2R}$. Since $\delta y(x)$ is usually a decimal, we perform linear interpolation in the Y direction to achieve this operation. After bending, some pixels will exceed the matrix with \emph{W} rows and \emph{L} columns. For these pixels out of the matrix, we directly discard them. For the blank part of the matrix, we fill it with pure white pixels. Now we have the curves as shown in \ref{imitation}(b). The purpose of the first bending operation is to make the head and tail of the lines sharp. The second time of bending is almost the same as the first time, but we preserve those pixels out of the matrix. The purpose of the second bending is to increase the curvature further. Now we have the curves as shown in \ref{imitation}(c). It is worth noting that these curves are essentially still straight lines, only looking more natural than straight lines. We model the strokes as such straight lines rather than arbitrarily shaped curves is because, on the one hand, the real pencil strokes are mostly like this, and on the other hand, the straight lines are convenient for our subsequent work.
\begin{figure}
  \includegraphics[width=\linewidth]{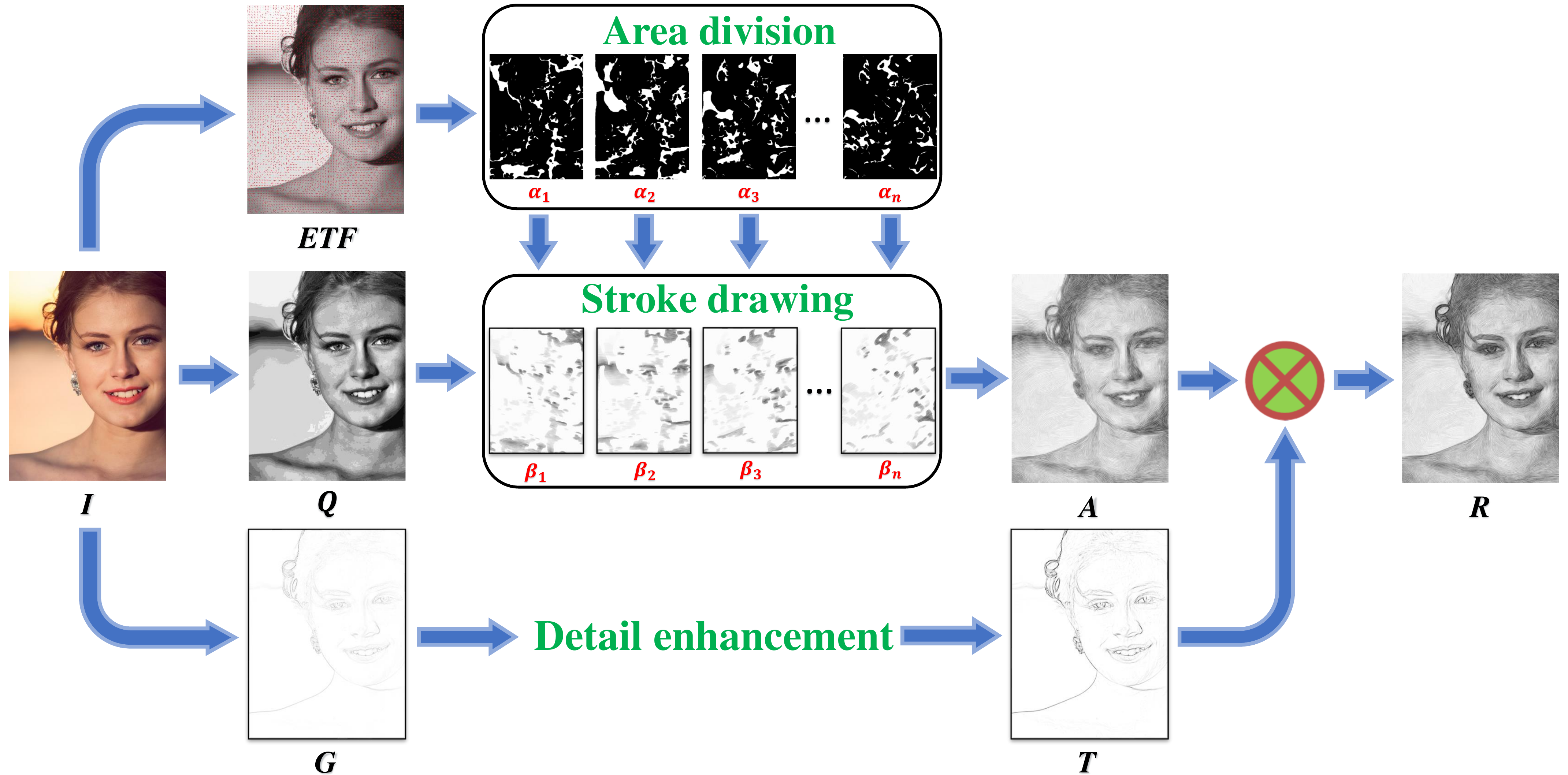}
  \caption{Schematic illustration of our algorithm. \emph{I} is the input. \emph{ETF} is the visualization of edge tangent flow vector field \cite{kang2007coherent}. \{$\alpha_{1}$, $\alpha_{2}$, \ldots, $\alpha_{n}$\} are the area divisions of the input according to the direction of \emph{ETF} vectors. \emph{Q} is the quantization result of \emph{I}. \{$\beta_{1}$, $\beta_{2}$, \ldots, $\beta_{n}$\} are the stroke drawing results of each area. \emph{A} is the aggregation of \{$\beta_{1}$, $\beta_{2}$, \ldots, $\beta_{n}$\}. \emph{G} is the gradient map of \emph{I}. \emph{T} is the edge map generated by \emph{G}. \emph{R} is the final result obtained by multiplying \emph{A} and \emph{T}.}

  \label{pipeline}
\end{figure}
\section{Guided Stroke Drawing}
Now we introduce how to do sketching by drawing one stroke every time on the canvas. To determine a stroke, we need to know the line's width \emph{W}, length \emph{L}, central pixel's gray value mean \emph{G}, starting point's coordinates, and this line's direction. For line width \emph{W}, now we specify the width of all strokes as a fixed value (we will discuss the influence of line width \emph{W} in Section 4.5 User Control). We will utilize the local characteristics of the input image to determine the other parameters.
\begin{figure*}
    \includegraphics[width=\linewidth]{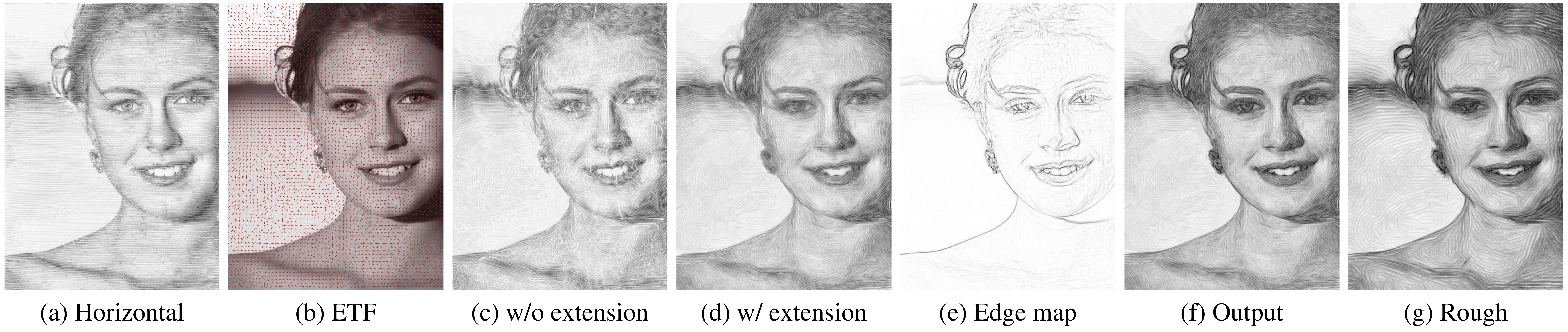}
	\caption{Some images used for algorithm introduction. (a) is the result of drawing lines only in the horizontal direction; (b) is the visualization of edge tangent flow vector field \cite{kang2007coherent}; (c) and (d) are the results of drawing lines without/with extension respectively; (e) is the edge map obtained by \cite{lu2012combining}; (f) is the drawing result of 5 pixel wide lines while lines in (g) are 9 pixel wide.} 
	\label{compare}
\end{figure*}
\subsection{Grayscale Guidance}
To reduce the difficulty of the task, in this section, we do not consider how to determine the direction of the strokes temporarily, but only show how to draw strokes in a fixed direction. Figure \ref{compare}(a) is the result of drawing strokes only in the horizontal direction. In the following, we will introduce how to draw it.

We first adjust the histogram distribution of the input gray image to improve its hue. We use contrast limited adaptive histogram equalization (CLAHE) \cite{zuiderveld1994contrast} to enhance the contrast of the input. Next, we uniformly quantify the image into several gray levels. We denote their gray values as \{$G_{1}$, $G_{2}$, ..., $G_{n}$\}. These values will be used as the strokes' central pixel gray value mean \emph{G}. As shown in Figure \ref{pipeline}, \emph{I} is the input and \emph{Q} is the result after quantization. Then we use \emph{Q} to search and determine the strokes' parameters by scanning in the horizontal direction.

Take the first row of the pixels in \emph{Q} as an example. Search out all the intervals in the first row where the pixels' gray value is less than or equal to $G_{1}$. Use the starting point and length of these intervals as the starting point and length of the strokes to be drawn; use $G_{1}$ as the strokes' central pixel gray value mean \emph{G}. Then we can draw several strokes in the horizontal direction (width \emph{W} is specified in advance). Here we define a new random variable $D\sim N(W,1)$, which is used to generate the lines' pixel distance in the vertical direction. We have searched and drawn strokes in the first row. Every next time we move down \emph{D} rows in the vertical direction and repeat the same operation as the first row until reaching the bottom of the quantization image \emph{Q}. In this way,  we could draw all the strokes with central pixel gray value mean $G_{1}$. Then we draw strokes for \{$G_{2}$, ..., $G_{n}$\} in the same way. In this process, different strokes' coverage regions will overlap. The gray value of the pixels in the overlapped region is determined by the minimum (darkest). Now we can get the drawing result shown in Figure \ref{compare}(a). Although we only introduce the method of drawing strokes in the horizontal direction, different directions are equivalent: we can rotate the input image by an angle clockwise before drawing strokes in the horizontal direction. After the drawing is done, we rotate it back counterclockwise. In this way, we can draw strokes in every direction.

\subsection{Direction Guidance}
We have introduced how to draw strokes in a fixed direction. Now suppose we can divide the picture into several areas. Only strokes in the same area have the same direction. Then we can use the method in Section 4.1 Grayscale Guidance to draw strokes for every area according to the area division. 

By observing real pencil drawings, it is easy to find the direction of strokes is usually along the edges' tangent \cite{kelen1974leonardo}. Therefore, we hope to use the edges of objects to guide the direction of the strokes nearby. However, it is difficult to predict the structure of an object from a 2D image. Actually, we do not need accurate predictions but only an estimation. We could use the input image's gradient information to do this estimation because gradient and edges are often closely related. Under the natural light condition, the change of light's intensity at objects' edges is often more apparent than in flat areas. Therefore, the gradient vector field could offer suggestions for determining the direction of the strokes. We use the edge tangent flow (\emph{ETF}) proposed by \cite{kang2007coherent} to estimate the strokes' direction for each pixel. The constructing of \emph{ETF} is as follows: First, calculate the modulus and direction of the gradient vector for each pixel. Then rotate the directions of these vectors counterclockwise by 90 degrees. Adjust the vectors' direction iteratively so that the direction of the vectors with small modulus tends to the direction of the vectors with larger modulus nearby. Finally, the direction of all the vectors will be roughly parallel to the tangent of the edges. The visualization of the edge tangent flow vector field is shown in Figure \ref{compare}(b). The small red arrows point to the direction of the \emph{ETF} vectors.

Now we could divide the input image into several areas according to the \emph{ETF}. We uniformly quantify the direction $0 \sim 2 \pi$ into \emph{n} values. Vectors with a phase difference of $\pi$ are regarded as the same direction. After quantifying the direction, pixels with the same direction are divided into one area. As shown in the “Area division" box in Figure \ref{pipeline}, \{$\alpha_{1}$, $\alpha_{2}$, \ldots, $\alpha_{n}$\} indicate the area divisions (pixels belong to a certain area are white, while the others are black). After drawing strokes with different directions for each area we can get \emph {n} results, indicated by \{$\beta_{1}$, $\beta_{2}$, \ldots, $\beta_{n}$\} in the “Stroke drawing" box in Figure \ref{pipeline}. 

\subsection{Area Merging and Detail Enhancement}
Now we aggregate \{$\beta_{1}$, $\beta_{2}$, \ldots, $\beta_{n}$\} into one picture, as shown in Figure \ref{compare}(c). There are obvious defects at the boundaries of different areas. Besides, the area division will cause a large number of very short strokes, which look like noise points. These two problems are solved by extending the head and tail of all strokes by \emph {2W} pixels (\emph {W} is the strokes' width), as shown in Figure \ref{compare}(d). \emph{A} in Figure \ref{pipeline} indicates the aggregation result of strokes in different directions, which could be obtained as this equation:
\begin{equation}
    A=minimum({ \beta_{1}}, {\beta_{2}}, \ldots,{\beta_{n}} )
\end{equation}
Now the strokes in different directions merge well and appear more continuous. However, extending the strokes also causes loss of detail clarity. For example, the lady's teeth and eyes in Figure \ref{compare}(d) are very unclear. Therefore, we need to enhance the details of the sketch. As shown in Figure \ref{pipeline}, \emph{G} is the gradient map of the input image. \emph{T} is the edge map obtained by \emph{G}. There are countless algorithms for generating the edge map from the gradient map. Here we adopt the linear convolution method of \cite{lu2012combining} because the edge map obtained by his method looks more like the result of a pencil drawing than other algorithms, as shown in Figure \ref{compare}(e). 

Finally, we multiply the edge map \emph{T} and the drawing result \emph{A} to get the final output \emph{R}, expressed as $R=A\cdot T$. The final output \emph{R} is shown in Figure \ref{compare}(f).

\subsection{Process Reconstruction}
The strokes' search method has been introduced: we first draw strokes in different directions for each area and then integrate them into the final output, which seems odd. However, due to all the strokes' parameters being determined and recordable when searching for, we can rearrange the strokes' drawing sequence in a more realistic and meaningful order. We call this “Process Reconstruction". When artists draw sketches, they usually draw the outlines first (which are often long or dark lines), and then the details (these lines are often short and not very dark). To imitate this, we use the index \emph{S} to measure whether each line is more likely to be an outline or a detail:
\begin{equation}
   S=(255-G)\times\sum\nolimits_{i \in D}T_{i}
\end{equation}
where $G$ is the stroke's central pixel gray value mean, $D$ is the set of pixels where the stroke covers, $T_i$ indicates the absolute value of gradient at pixel $i$. The larger the $S$, the greater the possibility that the stroke to be an outline. So we reconstruct the strokes' drawing process according to $S$ in descending order. Three examples are shown in Figure \ref{process}. It can be seen that when we only draw about 20\% of the total strokes, we can almost present the drawn objects. The whole drawing process (video demo) and more interesting results can be found in the supplementary material.

\subsection{User Control}

\subsubsection{Fineness}
Our pencil drawing's fineness is controlled by the strokes' width and the number of quantization order. Our method of searching strokes is essentially doing sampling, and the width \emph{W} of the strokes is the sampling interval. The wider the stroke, the lower the sampling frequency. Therefore, the wider the strokes, the rougher the pencil drawing will be. Figure \ref{compare}(g) is the result of $W=9$ while $W=5$ in Figure \ref{compare}(f). The quantization order's influence is the same as the general case: the larger the quantization order is, the less obvious the block effect is. Usually, we fix \emph{W} to be 5 pixels to get finer details. We fix the quantization order of direction to be $10$ because too few directions will make the texture fluency worse, but too many directions will not improve the visual experience. The quantization order of gray level could be chosen from 8-16 according to the input. Inputs with more low-frequency components require more quantization orders. More detailed explanation of
the hyperparameters’ influence on the drawing result could be found in the supplementary material.

\subsubsection{Color space conversion}
All of the above work is performed on the gray image, and only need to do color space conversion to achieve coloring.
We convert the original image to the YUV color space, replace the Y channel with the gray output, and then transfer back to RGB color space to obtain a colored pencil drawing. This coloring method is the same as \cite{lu2012combining}. The second column from the right in Figure \ref{process} shows our RGB result.

\begin{figure}
    \includegraphics[width=\linewidth]{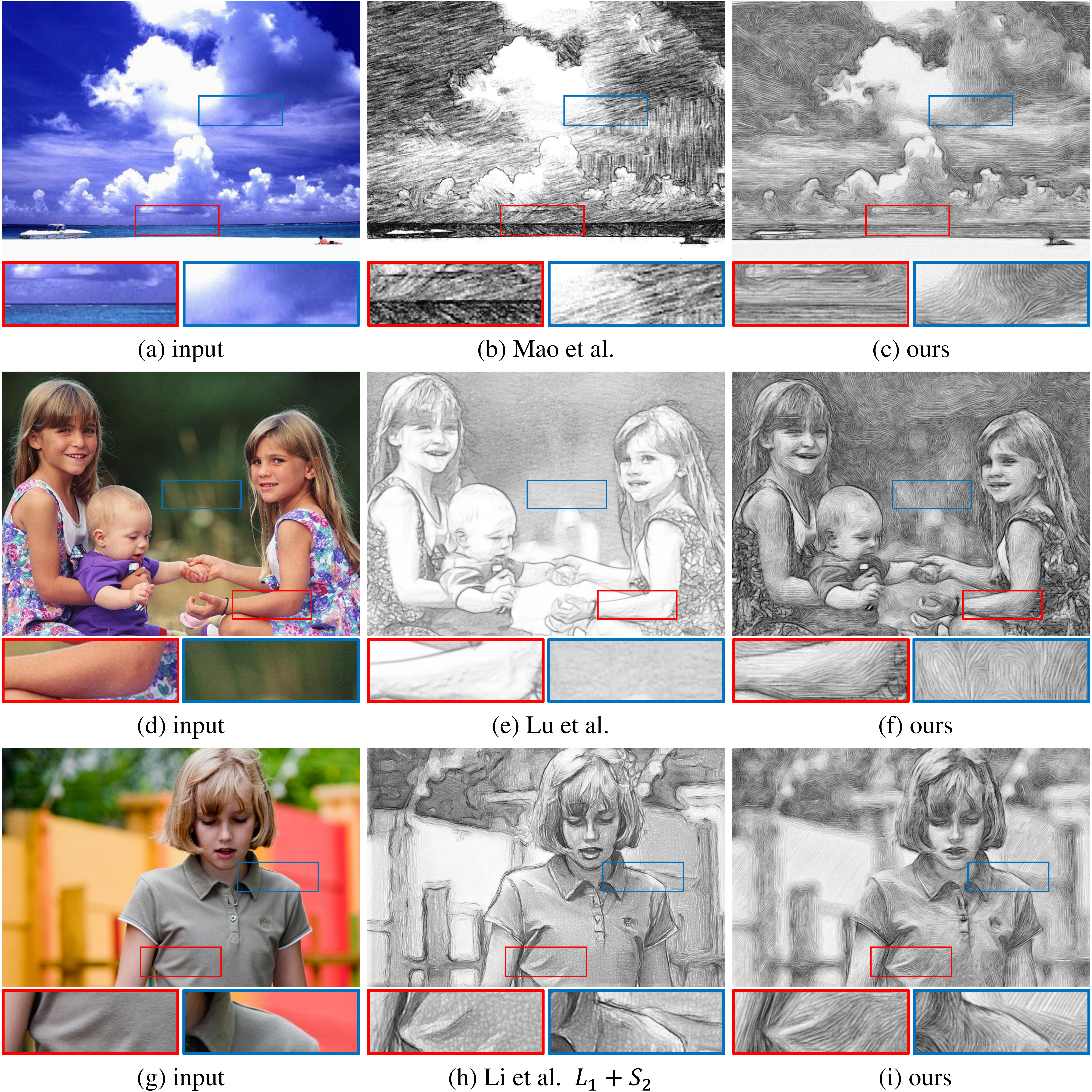}
	\caption{Comparison with several existing algorithms} 
	\label{C}
\end{figure}
\section{Experimental Results}
In this section, we compare our results with several representative methods to prove the effectiveness of our algorithm. The pictures being compared are all from the original paper. Due to the page limit, we only compared with three methods in this article. More comparisons and comparisons with more methods (especially neural methods) can be found in the supplementary material.



Firstly, we compare our proposed method with a classic LIC-based method. In the first row of Figure \ref{C}, (a) is the input, (b) is the result of \cite{mao2002automatic}, (c) is our result. Since LIC image is obtained by low-pass filtering a white noise input image, (b) introduces too much noise on the original image and looks dirty. The texture direction of (b) is very dull, with only three directions (30\degree, 90\degree, 135\degree), and the change of texture direction has nothing to do with semantic information. In the zoomed-in area (red border), the sea's texture direction in (b) is 135\degree while (c)'s is 0\degree, which is more in line with the texture of sea ripples. In the zoomed-in area (blue border), the texture direction of (b) is rigidly fixed at 30\degree  while (c)'s texture looks very fluent. (c) well expresses the feeling of clouds floating in the wind, and the overall visual effect is much cleaner and clearer than (b). The comparison shows that our method's result performs better in terms of aesthetic perception and is closer to the artists' technique.

Since we use \cite{lu2012combining}'s method to extract edges and enhance details, we make a comparison with \cite{lu2012combining} in the second row of Figure \ref{C}. \cite{lu2012combining}'s method can well depict the input's contours and edges, but their texture looks too smooth, more like adding a little noise to the gray image. Besides, \cite{lu2012combining}'s method uses histogram matching to fix the pencil drawing's tone, making many areas appear too pale, such as the girls' arms and the baby's head in (e), with only contours but no texture. In our result (f), the girls' arm and facial skin's texture direction is highly consistent with the actual structure of the human muscles, which is very aesthetic and vivid. The background in our result (f) has a strong style, while (e) lacks texture, very close to a gray image.

Now we compare with \cite{li2019im2pencil}, which is state of the art among neural methods. As shown in the third row of Figure \ref{C}, (g) is the input images, (h) is the result of \cite{li2019im2pencil}. Li et al.'s method can produce different kinds of style combinations. For example, $L_1$+$S_2$ means using the first type of outline and the second type of shading. So we just chose one of their style combinations that archives relatively good effects for comparison. Observe (h), it is not difficult to find apparent artifacts along the edges and borders, such as the girl's shoulder in the zoomed-in area (blue border). These artifacts deteriorate the fineness of their pencil drawing. (i) is our result, our method can preserve objects' contours clearly and portray tiny details. In terms of texture, it can be seen from the zoomed-in area (red border) that (h)'s texture cannot establish a direct connection with semantics (folds of clothes), and the texture of (h) can only reflect the shade of the input but not the input's structure. Our method displays the shade and fold of the cloth well at the same time.

\begin{table}
\centering
\begin{tabular}{|c|c|c|c|c|c|}
\hline
\multirow{2}{*}{Group} & \multirow{2}{*}{Evaluation} & \multicolumn{4}{c|}{Score (Average)} \\ \cline{3-6} 
                       &                             & {[}1{]}  & {[}2{]} & {[}3{]} & {[}4{]} \\ \hline
\multirow{5}{*}{1st}   & Stroke/Texture              & 66.4     & 52.8    & 73.1    & 88.6    \\ \cline{2-6} 
                       & Tone/Contrast               & 78.1     & 72.6    & 84.5    & 87.7    \\ \cline{2-6} 
                       & Stereoscopy                 & 54.3     & 68.9    & 80.4    & 84.2    \\ \cline{2-6} 
                       & Authenticity                & 60.7     & 65.3    & 83.1    & 95.3    \\ \cline{2-6} 
                       & Overall                     & 68.5     & 74.1    & 83.2    & 92.7    \\ \hline
2nd                    & Overall                     & 72.0     & 80.4    & 84.2    & 87.3    \\ \hline
\end{tabular}
\caption{People in the 1st group were asked to rate the result in terms of Stroke/Texture, Tone/Contrast, Stereoscopy, Authenticity, and Overall Perception, while the 2nd group only need to rate the last term. [1], [2], [3], [4] represent the methods of \cite{mao2002automatic}, \cite{lu2012combining}, \cite{li2019im2pencil}, and ours respectively. It can be seen that our method has the highest score in the opinions of both group of people.}
\end{table}

\section{User study}
We investigated the preferences of two groups of people (100 people in each group) to different pencil drawing algorithms. People in the first group (including ten recognized artists in our city and ninety professors/lecturers/graduates in painting-related majors from several universities) have received professional training in painting, while those in the second group do not. We gave ten sets of pictures to every participant. Every set of pictures included an input and the corresponding result of four pencil drawing algorithms. Participants didn't know the drawing was from which method and the ordering of four methods were shuffled in every set.
For the first group of participants, we provided them with a series of subjective evaluation indicators and asked them to rate these indicators for each result; for the second group, we only asked them to rate every result based on their overall perception. Score range was limited in $0\sim 100$ and participants were asked to give distinguished scores. The feedback results are shown in Table 1.  

After completing the above survey, we used A, B, C, and D to anonymously represent these four algorithms and let each participant choose their favorite algorithm. We counted the number of people who voted for each algorithm. Then we showed our results' drawing process and informed participants that the other three methods couldn't generate process. Now we let the participants choose their favorite algorithm again. The survey results are shown in Table 2: most participants chose our method as their favorite; after watching our drawing process, more people voted for our method. 

\begin{table}
\centering
\begin{tabular}{|c|c|c|c|c|c|}
\hline
\multirow{2}{*}{Group} & \multirow{2}{*}{See Drawing Process} & \multicolumn{4}{c|}{User Preference} \\ \cline{3-6} 
                       &                                  & {[}1{]}  & {[}2{]}  & {[}3{]}  & {[}4{]}  \\ \hline
\multirow{2}{*}{1st}   & Before                           & 3        & 14       & 25       & 58       \\ \cline{2-6} 
                       & After                            & 2        & 9        & 12       & 77       \\ \hline
\multirow{2}{*}{2nd}   & Before                           & 2        & 19       & 27       & 52       \\ \cline{2-6} 
                       & After                            & 0        & 8        & 11       & 81       \\ \hline
\end{tabular}
\caption{User Preference indicates how many people voted for each algorithm. [1], [2], [3], [4] represent the methods of \cite{mao2002automatic}, \cite{lu2012combining}, \cite{li2019im2pencil}, and ours respectively.}
\end{table}

\section{Conclusions}
In our work, we statistically analyze and explain the texture of real pencil drawings and propose a controllable pencil stroke generation mechanism. On this basis, we implement an image-to-pencil automatic drawing algorithm: we use the edge tangent flow vector field to guide the direction of the strokes; use the gray image to determine the location, length, and shade of the strokes; use the edge map for detail enhancement. Our method is a mathematical procedural algorithm with good interpretability. Comparison with other pencil drawing algorithms shows our method outperforms in terms of texture quality, style, and user evaluation. Our most prominent advantage is that we have the drawing process.

\section{Acknowledgments}
This work was supported by National Science Foundation of China (U20B2072, 61976137, U1611461).
\bibliography{Tong}
\end{document}